# Dimensional Neuroimaging Endophenotypes: Neurobiological Representations of Disease Heterogeneity Through Machine Learning

Junhao Wen[1*], Mathilde Antoniades[2], Zhijian Yang[2], Gyujoon Hwang[3], Ioanna Skampardoni[2], Rongguang Wang[2], Christos Davatzikos[2,*]

[1]Laboratory of AI and Biomedical Science (LABS), Stevens Neuroimaging and Informatics Institute, Keck School of Medicine of USC, University of Southern California, Los Angeles, CA, USA
[2]Artificial Intelligence in Biomedical Imaging Laboratory (AIBIL), Center for AI and Data Science for Integrated Diagnostics (AI$^2$D), Perelman School of Medicine, University of Pennsylvania, Philadelphia, PA, USA
[3]Psychiatry and Behavioral Medicine, Medical College of Wisconsin, Watertown Plank Rd, Milwaukee, WI, USA

*Corresponding authors:

Junhao Wen, Ph.D. – junhaowe@usc.edu
2025 Zonal Ave, Los Angeles, CA 90033, United States
Christos Davatzikos, Ph.D. - christos.davatzikos@pennmedicine.upenn.edu
3700 Hamilton Walk, 7th Floor, Philadelphia, PA 19104, USA

# Abstract

Machine learning has been increasingly used to obtain individualized neuroimaging signatures for disease diagnosis, prognosis, and response to treatment in neuropsychiatric and neurodegenerative disorders. Therefore, it has contributed to a better understanding of disease heterogeneity by identifying disease subtypes that present significant differences in various brain phenotypic measures. In this review, we first present a systematic literature overview of studies using machine learning and multimodal MRI to unravel disease heterogeneity in various neuropsychiatric and neurodegenerative disorders, including Alzheimer's disease, schizophrenia, major depressive disorder, autism spectrum disorder, multiple sclerosis, as well as their potential in transdiagnostic settings. Subsequently, we summarize relevant machine learning methodologies and discuss an emerging paradigm which we call dimensional neuroimaging endophenotype (DNE). DNE dissects the neurobiological heterogeneity of neuropsychiatric and neurodegenerative disorders into a low-dimensional yet informative, quantitative brain phenotypic representation, serving as a robust intermediate phenotype (i.e., endophenotype) largely reflecting underlying genetics and etiology. Finally, we discuss the potential clinical implications of the current findings and envision future research avenues.

# 1. Main

Over the past two decades, magnetic resonance imaging (MRI) and machine learning have emerged as foundational tools and techniques for studying human brain aging and disease[1]. Researchers have proposed an array of individual-level imaging signatures[2–10,10–12] to quantify disease and aging effects using state-of-the-art machine learning techniques. However, disease heterogeneity poses a major obstacle to their potential clinical implementation. Disease heterogeneity can manifest in various aspects such as neuroanatomy and function[13–16], clinical symptoms[17], and genetics[18]. Critically, case-control studies largely overlooked such heterogeneity, leading to limited applicability due to the inability to capture diverse, multifaceted underlying biological processes that collectively give rise to the ultimate manifestation of clinical symptoms. Furthermore, it is anticipated that the heterogeneity in the underlying etiology and clinical manifestations thereof will also give rise to variability in response to experimental pharmacotherapeutics[19]. Therefore, the effectiveness of the drugs developed and tested in the '*1-for-all*' unitary group of patients, such as Alzheimer's disease (AD), may be hindered since the study population may represent a mixture of multiple pathological processes.

The research community has increasingly leveraged machine learning to address this challenge. A growing body of literature has focused on applying clustering methods to brain MRI data to derive disease subtypes. The term "subtype" delineates a predetermined clustering resolution with a fixed threshold (e.g., 0.5 for probability-based models), potentially disregarding the evolving nature of brain diseases across a spectrum. This approach assumes singular imaging patterns within patients, ignoring the potential for multiple presentations or dynamic changes over time. In this review, we conceptualize an emerging paradigm – Dimensional Neuroimaging Endophenotype (DNE) – to model and quantify the neurobiological heterogeneity of brain diseases, digitizing disease heterogeneity and allowing for the co-expression of multiple imaging patterns within the same patient. **Fig. 1** details the proposed DNE framework. These DNEs hold particular significance as intermediate phenotypes, akin to endophenotypes initially introduced in psychiatric genetics[20]. Situated within the causal trajectory of brain disorders, they bridge the gap between underlying genetic variants (such as single nucleotide polymorphisms) and the eventual manifestation of clinical symptoms (exo-phenotypes). Consequently, they emerge as useful instruments for investigating the origins and progression of brain diseases.

While previous reviews on this topic are available elsewhere[21,22,23], they often lack a systematic literature overview or focus on a single disease, such as AD. In response to the rapidly growing interest in unraveling disease heterogeneity using machine learning, the present study seeks to provide a comprehensive and systematic review of the state-of-the-art in several common neuropsychiatric and neurodegenerative disorders. In particular, this review entailed a rigorous bibliometric search aimed at identifying relevant research publications in AD, schizophrenia (SCZ), major depressive disorder (MDD), autism spectrum disorder (ASD), multiple sclerosis (MS), and transdiagnostic disorders (TD). **Table 1** presents the surveyed papers in our systematic review, and **Supplementary eMethod 1** details the inclusion criteria. Subsequently, we briefly overviewed commonly employed machine learning methodologies for disease heterogeneity and introduced the DNE framework through weakly-supervised clustering techniques. Finally, we thoroughly examined and deliberated upon the surveyed studies conducted within each brain condition and disease, delineating prospective paths for future research endeavors.

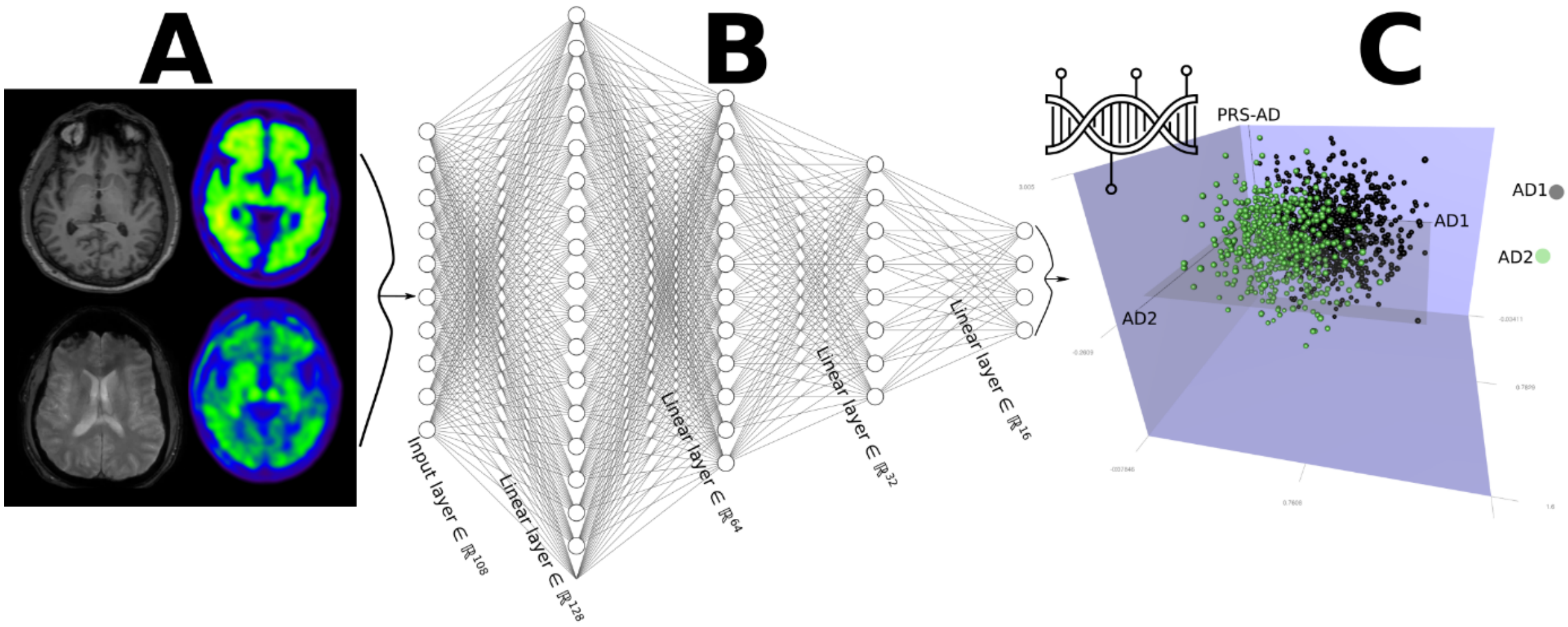

**Figure 1. The framework of Dimensional Neuroimaging Endophenotypes (DNE) to unravel the neurobiological heterogeneity of brain diseases**. **A**) Imaging-derived phenotypes (IDPs) are extracted from multimodal MRI, including T1-weighted MRI, T2-weighted MRI, and PET. **B**) Machine learning models are applied to IDPs to position patients into multiple (*k*) DNEs (e.g., AD1 and AD2). **C**) As such, they represent a reliable instrument for re-evaluating disease-related hypotheses and identifying suitable populations for drug development. Additional information, such as PRS, can be integrated into this low-dimensional latent space. Abbreviation: Alzheimer's disease: AD; polygenic risk scores: PRS.

**Table 1**. Surveyed studies using machine learning to dissect the neurobiological heterogeneity of Alzheimer's disease (AD), schizophrenia (SCZ), major depressive disorder (MDD), autism spectrum disorder (ASD), multiple sclerosis (MS), and transdiagnostic disorders (TD). Abbreviations: BD: bipolar disorder; ROP: recent-onset psychosis; ROD: recent-onset depression; CN: healthy control; T1w MRI: T1-weighted MRI; dMRI: diffusion MRI; fMRI: functional MRI; FES: first-episode schizophrenia; VBM: voxel-based morphology; GMM: Gaussian mixture model; MWF: myelin water fraction; NDI: neurite density index; CIS: Clinically Isolated Syndrome; ALFF: amplitude of low-frequency fluctuations; DL: deep learning; DNN: default mode network; VAN: ventral affective network; NMF: non-negative matrix factorization; CNN: convolutional neural network; LDA: Latent Dirichlet allocation; ID: Internalizing disorders. We included a "**Code Availability**" section that provides access to the software used in the machine learning methodologies whenever available. This systematic review encompasses papers published from January 1990 to January 15, 2023 (**Supplementary eMethod 1**). Recognizing the rapidly evolving nature of the field, we have made **Table 1** publicly accessible at the following link: https://docs.google.com/spreadsheets/d/1KGA9pyQsxcsshxp70gICt5H-7Wt73bf03NI_UnyOuTg/edit#gid=0. We encourage the research community to contribute additional published studies on this topic.

| Study | Modality | Sample size | Method | Category | Subtype characteristics |
|---|---|---|---|---|---|
| AD | | | | | |
| Yang et al., 2021[14] | T1w MRI | 1620 CN, 1212 MCI/AD | Smile-GAN | Semi-supervised | Subtype1 shows preserved brain volume; Subtype2 shows mild diffuse atrophy; Subtype3 shows focal medial temporal lobe atrophy; Subtype4 shows severe atrophy over the whole brain. |
| Duong et al., 2022[24] | T1w MRI, PET | 289 MCI/AD | Hierarchical clustering | Unsupervised | Relative to neurodegeneration, Subtype1 shows high cortical resilience to tau; Subtype2 shows limbic resilience to tau; Subtype3 shows low cortical resilience to tau; Subtype4 shows consistent/canonical neurodegeneration and tau pathologies; Subtype5 shows cortical susceptibility to tau; Subtype6 shows limbic susceptibility to tau. |
| Vogel et al., 2021[25] | PET | 1143 CN/MCI/AD | SuStaIn | Unsupervised | Subtype1 shows a limbic-predominant tau pattern; Subtype2 represents a medial temporal lobe-sparing pattern; Subtype3 shows a posterior tau pattern; Subtype4 shows a lateral temporal tau pattern. |
| Poulakis et al., 2020[26] | T1w MRI | 31 CN, 72 AD | Multivariate Mixture of Generalized Mixed effect Models | Unsupervised | Subtype1 shows typical diffuse atrophy pattern; Subtype2 shows minimal atrophy patterns; Subtype3 shows hippocampal sparing atrophy patterns. |
| Chen et al., 2022[27] | fMRI | 373 CN, 350 MCI, 377 AD | NMF | Unsupervised | Subtype1 shows diffuse and mild functional connectivity disruption; Subtype2 shows predominantly decreased connectivity in the default mode network accompanied by an increase in the prefrontal circuit; Subtype3 shows predominantly decreased connectivity in the anterior cingulate cortex accompanied by an increase in prefrontal cortex connectivity; Subtype4 shows predominantly decreased connectivity in the basal ganglia accompanied by an increase in prefrontal cortex connectivity. |
| Wen et al., 2022[28] | T1w MRI | 541 CN, 848 MCI, 339 AD | MAGIC | Semi-supervised | Subtype1 shows focal atrophy in temporal regions; Subtype2 shows whole-brain atrophy with the most severe atrophy in temporal and hippocampus regions; |
| | | | | | Subtype3 shows atypical AD patterns without affecting the hippocampus and temporal lobes. Subtype4 shows preserved relatively normal brain anatomy. |
| Young et al., 2018[29] | T1w MRI | 349 CN, 734 MCI, 227 AD | SuStaIn | Unsupervised | Subtype1 has atrophy starting in the hippocampus and amygdala; Subtype2 has atrophy starting in the nucleus accumbens, insula, and cingulate; Subtype3 has atrophy starting in the pallidum, putamen, nucleus accumbens, and caudate. |
| Dong et al., 2017[30] | T1w MRI | 399 CN, 510 MCI, 314 AD | CHIMERA | Semi-supervised | Subtype1 shows the least amount and extent of atrophy; Subtype2 shows widespread but relatively most severe temporal atrophy; Subtype3 shows a more diffuse atrophy pattern; Subtype4 shows moderate localized atrophy in the hippocampus and the anterior-medial temporal cortex. |
| Varol et al., 2017[31] | T1w MRI | 177 CN, 123 AD | HYDRA | Semi-supervised | Subtype1 shows diffuse atrophy patterns; Subtype2 shows atrophy in the bilateral parietal lobe, bilateral temporal cortex, bilateral dorsolateral frontal lobe, precuneus; Subtype3 shows atrophy predominantly in the bilateral medial temporal cortex. |
| Corriveau-Lecavalier et al., 2023[32] | PET | 52 AD | Hierarchical clustering | Unsupervised | Subtype1 shows hypometabolism in hetero-modal cortices of the right hemisphere; Subtype2 shows an overall milder pattern of hypometabolism mostly concentrated in parietal areas bilaterally; Subtype3 shows hypometabolism in hetero-modal cortices of the left hemisphere; Subtype4 shows hypometabolism in hetero-modal cortices bilaterally. |
| Kwak et al., 2021[33] | T1w MRI | 109 CN, 380 MCI, 110 AD | CNN | Supervised | Subtype1 shows CN-like brain atrophy patterns; Subtype2 shows AD-like brain atrophy patterns. |
| Zhang et al., 2016[34] | T1w MRI | 188 AD | Bayesian LDA | Unsupervised | Subtype1 shows atrophy throughout the cortex; Subtype2 shows extensive atrophy in the medial temporal lobe; Subtype3 shows atrophy in the cerebellum, striatum, and thalamus |
| Filipovych et al., 2012[35] | T1w MRI | 126 CN, 17 MCI | JointMMCC | Semi-supervised | Subtype1 shows atrophy in several temporal, parietal, occipital and temporal and medial cortical regions; Subtype2 shows relatively normal brain anatomy. |
| Lee et al., 2021[36] | PET | 37 CN, 60 MCI | Louvain method | Unsupervised | Subtype1 shows greater THK5351 retention in the limbic regions; Subtype2 shows greater THK5351 retention in diffuse brain regions; Subtype3 shows no significantly greater THK5351 retention; Subtype4 shows greater THK5351 retention in AD-like brain regions. |
| Sun et al., 2023[37] | PET | 247 CN, 301 MCI | SuStaIn | Unsupervised | For Subtype1, amyloid accumulates sequentially in subcortical regions, cingulate, insula, and then cortical areas; For Subtype2, amyloid accumulates sequentially in cingulate, cortical regions, insula, and then the subcortical regions. |
| Noh et al., 2014[38] | T1w MRI | 152 AD | Hierarchical clustering | Unsupervised | Subtype1 shows medial temporal–dominant atrophy; Subtype2 shows parietal-dominant atrophy; Subtype3 shows diffuse atrophy pattern in which nearly all association cortices revealed atrophy. |
| Toledo et al., 2022[39] | T1w MRI, PET | 214 CN, 282 MCI/AD | Robust Collaborative Clustering | Unsupervised | Subtype1 shows greater atrophy in limbic regions; Subtype2 shows diffuse atrophy in the parietal–occipital–temporal circuit; Subtype3 shows hippocampal-sparing atrophy but also involves other diffuse brain regions. |
| Jeon et al., 2019[40] | T1w MRI, PET | 60 CN, 83 AD | Hierarchical Clustering | Unsupervised | Subtype1 shows a medial temporal-dominant subtype; Subtype2 shows a parietal-dominant subtype; Subtype3 shows a diffuse atrophy subtype. |
| SCZ | | | | | |
| Arnedo et al., 2015[41] | dMRI | 47 SCZ, 36 CN | Generalized factorization method | Unsupervised | Subtype1 shows low FA in the genu of the corpus callosum; Subtype2 shows low FA in the fornix and external capsule; Subtype3 shows low FA in the splenium of the corpus callosum, retro-lenticular limb, and posterior limb of the internal capsule; Subtype4 shows low FA in the anterior limb of the internal capsule. |
| Chand et al., 2020[42] | T1w MRI | 307 SCZ, 364 CN | HYDRA | Sem-supervised | Subtype1 shows widespread atrophy, including the thalamus, medial temporal, and medial prefrontal cortex; Subtype2 shows larger volumes in basal ganglia. |
| Dwyer et al., 2018[43] | T1w MRI | 71 SCZ, 74 CN | Fuzzy c-means algorithm | Unsupervised | Subtype1 shows the involvement of the insula, medial frontal, temporal, and parietal lobes; Subtype2 shows more diffuse patterns associated with the medial frontal, lateral frontal, and temporal cortex. |
| Honnorat et al., 2019[44] | T1w MRI | 157 SCZ, 169 CN | CHIMERA | Sem-supervised | Subtype1 shows brain regions in temporal-thalamic-peri-Sylvian; Subtype2 shows frontal regions and the thalamus; Subtype3 shows a mixed pattern of Subtype1 and Subtype2. |
| Liang et al., 2021[45] | fMRI | 300 SCZ, 169 CN | Spectral clustering | Unsupervised | Compared to CN, Subtype1 shows the opposite direction of FC between the ventromedial prefrontal cortex and anterior cingulate cortex, and the same direction of FC between the ventromedial prefrontal cortex and right posterior parietal cortex; Subtype2 exhibits the same direction of FC between the ventromedial prefrontal cortex and anterior cingulate cortex, and right posterior parietal cortex. |
| Pan et al., 2020[46] | T1w MRI | 179 SCZ, 77 CN | K-means | Unsupervised | Subtype1 shows global cortical thickness reduction; Subtype2 shows an intact brain; Subtype3 shows thickness reduction in the lingual, inferior parietal, lateral occipital lobes, and insula. |
| Shi et al., 2022[47] | T1w MRI | 534 SCZ, 521 CN | K-means | Unsupervised | Subtype1 shows moderate deficits of subcortical nuclei and enlarged striatum and cerebellum; Subtype2 displays cerebellar atrophy and more severe subcortical nuclei atrophy. |
| Sugihara et al., 2017[48] | T1w MRI | 108 SCZ, 121 CN | K-means | Unsupervised | There was substantial overlap between the patterns of cortical thickness in all 6 Subtype. Subtype1 exhibited the most extensive cortical thinning, particularly in the medial prefrontal and temporal regions, while the other 5 Subtype exhibited reduced cortical thickness in the medial frontal or temporal lobe. |
| Wen et al., 2022[49] | T1w MRI | 583 SCZ, 583 CN | MAGIC | Sem-supervised | Subtype1: enlarged striatum; Subtype2: diffuse brain atrophy over the entire brain. VBM results were also shown for other clustering solutions (i.e., 3 and 4 Subtype solutions). |
| Xiao et al., 2022[50] | T1w MRI | 299 SCZ, 403 CN | Density peak-based clustering | Unsupervised | FES patients in Subtype1 show decreased surface area, thickness, and volume, mainly in cortical-thalamic-cortical circuitry, and increased thickness in the left rostral anterior cingulate gyrus, while FES patients in Subtype2 and Subtype3 show no significant cortical or subcortical alteration; In midcourse schizophrenia patients, Subtype1 patients show widespread gray matter deficits in all lobes and the insular cortex and bilateral hippocampus while showing increased gray matter volume in bilateral pallidum. Subtype2 shows decreased gray matter volume in the left hippocampus. Subtype3 shows no significant brain alteration. |
| Zhao et al., 2022[51] | T1w MRI | 194 SCZ, 290 CN | K-means | Unsupervised | Subtype1 shows widespread neuroanatomic changes relative to controls, affecting all subcortical and multiple regional cortical volumes; Subtype2 displays significantly increased volume in the bilateral pallidum and limited cortical deficits. |
| MDD | | | | | |
| Drysdale et al., 2017[52] | fMRI | 458 MDD, 730 CN | Hierarchical clustering | Unsupervised | Reduced FC in front-amygdala networks is presented in Subtype1 and Subtype4; Hyperconnectivity in thalamic and frontostriatal networks is pronounced in Subtype3 and Subtype4; Reduced FC in cingulate and orbitofrontal areas is displayed in Subtype1 and Subtype2. |
| Price et al., 2017[53] | fMRI | 80 MDD | Walktrap | Unsupervised | Subtype1 uniquely shows pregenual anterior cingulate cortex to posterior cingulate cortex and dorsal anterior cingulate cortex to right insula functional connectivity path; Subtype2 uniquely shows dorsal anterior cingulate cortex to right parietal and left insula to right amygdala FC path. |
| Tokuda et al., 2018[54] | fMRI | 67 MDD, 67 CN | Multiple co-clustering | Unsupervised | Subtype1 shows both increased and decreased FC in DMN and ECN; Subtype2 shows primarily decreased FC in these networks; Subtype3 shows a smaller extent, compared to Subtype1, increased and decreased FC in these networks. |
| Liang et al., 2020[55] | fMRI | 690 MDD, 707 CN | K-means | Unsupervised | Subtype1 shows decreased functional connectivity in DNN; Subtype2 shows increased functional connectivity in DNN, including the left superior frontal cortex and left precuneus cortex, left superior frontal cortex, and left posterior cingulate cortex, and left superior frontal cortex and right ventral medial prefrontal cortex. |
| Woody et al., 2021[56] | fMRI | 70 MDD | Walktrap | Unsupervised | Subtype1 shows the fewest activated FC pathways; Subtype2 shows unique bidirectional VAN/DMN negative feedback; Subtype3 shows the most activated FC pathways.. |
| Wen et al., 2022[57] | T1w MRI | 501 LLD, 495 CN | HYDRA | Semi-supervised | Subtype1 shows larger gray matter volume in bilateral thalamus, putamen, and caudate; |
| | | | | | Subtype2 shows decreased gray matter volume in bilateral anterior and posterior cingulate gyri, superior, middle, and inferior frontal gyri, gyrus rectus, insular cortices, superior, middle, and inferior temporal gyri. |
| ASD | | | | | |
| Hrdlicka et al., 2005[58] | T1w MRI | 64 ASD | Hierarchical clustering | Unsupervised | Subtype1 shows the largest corpus callosum; Subtype2 shows the largest amygdala and hippocampus; Subtype3 shows the largest nucleus caudate and smallest hippocampus; Subtype4 shows the smallest corpus callosum, amygdala, and nucleus caudate |
| Hong et al., 2018[59] | T1w MRI | 107 ASD, 113 CN | Hierarchical clustering | Unsupervised | Subtype1 shows cortical thickening, increased surface area, tissue blurring; Subtype2 shows cortical thinning, decreased distance; Subtype3 shows increased distance |
| Chen et al., 2019[60] | T1w MRI | 356 ASD, 425 CN | K-means | Unsupervised | Subtype1 shows decreased prefrontal gray matter volume; Subtype2 shows increased temporal lobe volume, while decreased prefrontal and occipital volume; Subtype3 shows increased temporal lobe volume |
| Easson et al., 2019[61] | fMRI | 145 ASD, 121 CN | K-means | Unsupervised | Subtype1 shows greater FC between networks, particularly between the default mode network and the others; Subtype2 shows greater FC within networks |
| Jao Keehn et al., 2019[62] | fMRI | 57 ASD, 51 CN | K-means | Unsupervised | Subtype1 shows greater occipitofrontal FC; Subtype2 shows weaker occipitofrontal FC |
| Tang et al., 2020[63] | fMRI | 306 ASD | Bayesian | Unsupervised | Subtype1 shows weaker FC within and between perceptual-motor networks, and greater FC between perceptual-motor and association networks, and between somatic motor and subcortical regions; Subtype2 shows opposite patterns to Subtype1, with subtle deviations such as greater FC within default mode networks; Subtype3 shows greater FC between visual and somatomotor networks, and weaker FC within default mode and visual networks. |
| Aglinskas et al., 2022[64] | T1w MRI | 470 ASD, 512 CN | Gaussian mixture | Semi-supervised | No distinct Subtype: ASD-related neuroanatomical variation is better captured by continuous dimensions rather than by discrete categories. |
| Liu et al., 2022[65] | T1w MRI | 221 ASD, 257 CN | HYDRA | Semi-supervised | Subtype1: widespread brain volume increase; Subtype2: widespread brain volume decrease |
| Shan et al., 2022[66] | T1w MRI | 496 ASD, 560 CN | Gaussian mixture | Unsupervised | Subtype1 shows larger gray matter volume; Subtype2 shows decreased gray matter volume: Subtype3 shows largest gray matter volume |
| Hwang et al., 2023[67] | T1w MRI | 307 ASD, 362 CN | HYDRA | Semi-supervised | Subtype1 shows widespread brain volume decrease except for the orbital part of the inferior frontal gyrus; Subtype2 shows larger subcortical structures, especially pallidum and internal capsule; Subtype3 shows larger frontal gray matter and insula |
| MS | | | | | |
| Eshaghi et al., 2021[68] | T1-weighted, T2-weighted, T2-FLAIR | 6322 MS | SuStaIn | Unsupervised | Subtype1, 2, and 3 display cortex-led (cortical atrophy in the occipital, parietal, and frontal cortex), normal-appearing white |
| | MRI | | | | matter-led (a reduction in T1/T2 ratio of the cingulate bundle and corpus callosum), and lesion-led (early and extensive accumulation of lesions) characteristics, respectively. |
| Pontillo et al., 2022[69] | T1-weighted, and T2-FLAIR MRI | 425 MS, 148 CN | SuStaIn | Unsupervised | Subtype1 is characterized by the initial volume loss of subcortical gray matter structures followed by lesion accrual and cortical atrophy; Subtype2 shows cortical volume loss preceding DGM atrophy and lesion accumulation |
| Crimi et al., 2014[70] | T1 MRI | 25 MS | Spectral clustering | Unsupervised | Subtype1 comprises lesions of different dimensions (small, medium, large) and is generally gadolinium (Gd)-enhanced only; Subtype2 shows relatively medium, and large lesions and is with co-presence of ringing USPIO and focal Gd enhancement; Subtype3 comprises relatively medium lesions present and non-focal USPIO and Gd enhancement. |
| TD | | | | | |
| Chang et al., 2021[71] | fMRI | 581 SCZ, MDD, and BD, 363 CN | Unknown | Unknown | Subtype1 represents an archetypal dimension that ALFF is significantly increased in frontal areas and significantly decreased in posterior areas; DE2 is an atypical dimension that ALFF is significantly decreased in frontal areas, and significantly increased in posterior areas |
| Laousis et al., 2022[72] | T1w MRI | 155 ROP, 147 ROD 275 CN | HYDRA | Sem-supervised | Subtype1 has widespread gray matter volume deficits and more positive, negative, and functional deficits (impaired cluster); Subtype2 reveals a more preserved neuroanatomical signature and more core depressive symptomatology (preserved cluster). |
| Planchuelo-Gómez et al., 2020[73] | T1w MRI | 61 SCZ, 28 BD, 50 CN | K-means | Unsupervised | Subtype1 shows decreased cortical thickness and area values, as well as lower subcortical volumes and higher cortical curvature in some regions, as compared to Subtype2. |
| Kaczkurkin et al., 2020[74] | T1w MRI | 715 ID, 426 CN | HYDRA | Semi-supervised | Subtype1 shows smaller brain volumes and reduced cortical thickness; Subtype2 shows greater volume and cortical thickness. |

## 2. Machine learning methodology and dimensional neuroimaging endophenotypes

In recent years, significant strides have been made in the evolution of pioneering machine learning methodologies to tackle disease heterogeneity. These advancements broadly fall into two primary categories: *i*) unsupervised methodologies, which encompass techniques that do not rely on labeled data for training, and *ii*) weakly-supervised clustering[75], a subset of methods that leverage a combination of reference and target data to delineate distinct patterns within disease populations.

Initial attempts to address this issue utilized unsupervised clustering algorithms, like K-means, specifically tailored to imaging-derived features such as regions of interest (ROIs) from T1w MRI scans. These algorithms operate directly within the patient domain, organizing patients into clusters based solely on similarities or differences derived from their data (**Fig. 2A**). One advanced method, SuStaIn[29], is designed for subtype and stage inference, conceptualizing subjects exhibiting a particular biomarker progression pattern as a subtype. SuStaIn models the evolution of biomarkers within each subtype using a linear z-score model, an extended version of the original event-based model[76], where each biomarker follows a piecewise linear trajectory over a shared timeframe. Its key advantage lies in its ability to analyze purely cross-sectional data while providing estimates of imaging signatures for subtypes and stages.

Recent efforts have introduced weakly-supervised clustering methodologies aimed at establishing a "*1-to-k*" mapping between the healthy control (CN) and patient (PT) domains (depicted in **Fig. 2B**) to effectively model the underlying progression and course of disease[14,49,57,67,72,77]. Weakly-supervised clustering methods analyze the nuanced heterogeneity by aiming to extract data-driven and neurobiologically plausible subtypes. Their fundamental approach involves seeking a "*1-to-k*" mapping between the reference CN group and the PT group, specifically identifying clusters shaped by distinct pathological trajectories rather than relying solely on overall similarities or differences in data, as is typical in traditional unsupervised clustering methods.

These models were primarily inspired by the idea of subtypes, presuming that patients would be assigned a categorical phenotype representing a single, distinct imaging pattern. However, this assumption might not be biologically true, considering that brain diseases typically progress along a continuum and might manifest varying degrees of multiple imaging atrophy patterns. Therefore, this review conceptualized the DNE framework that models disease heterogeneity as a quantitative phenotype that can co-exist within the same patient for multiple dimensions. In a recent investigation, our group introduced the Surreal-GAN model to unravel the neuroanatomical diversity within AD) and found two DNEs (R1 and R2)[3]. Our findings demonstrated the suitability of these dimensions for subsequent genome-wide associations due to their adherence to a normal distribution, circumventing the collinearity issue commonly encountered in probability-based models where probabilities must sum up to 1. Crucially, the DNE framework posits an association between these DNEs and underlying genetics, as illustrated in **Fig. 1C**. This supports the well-established endophenotype hypothesis[20,78], corroborated by findings from our recent studies by linking these DNEs with common SNPs[4,6,67].

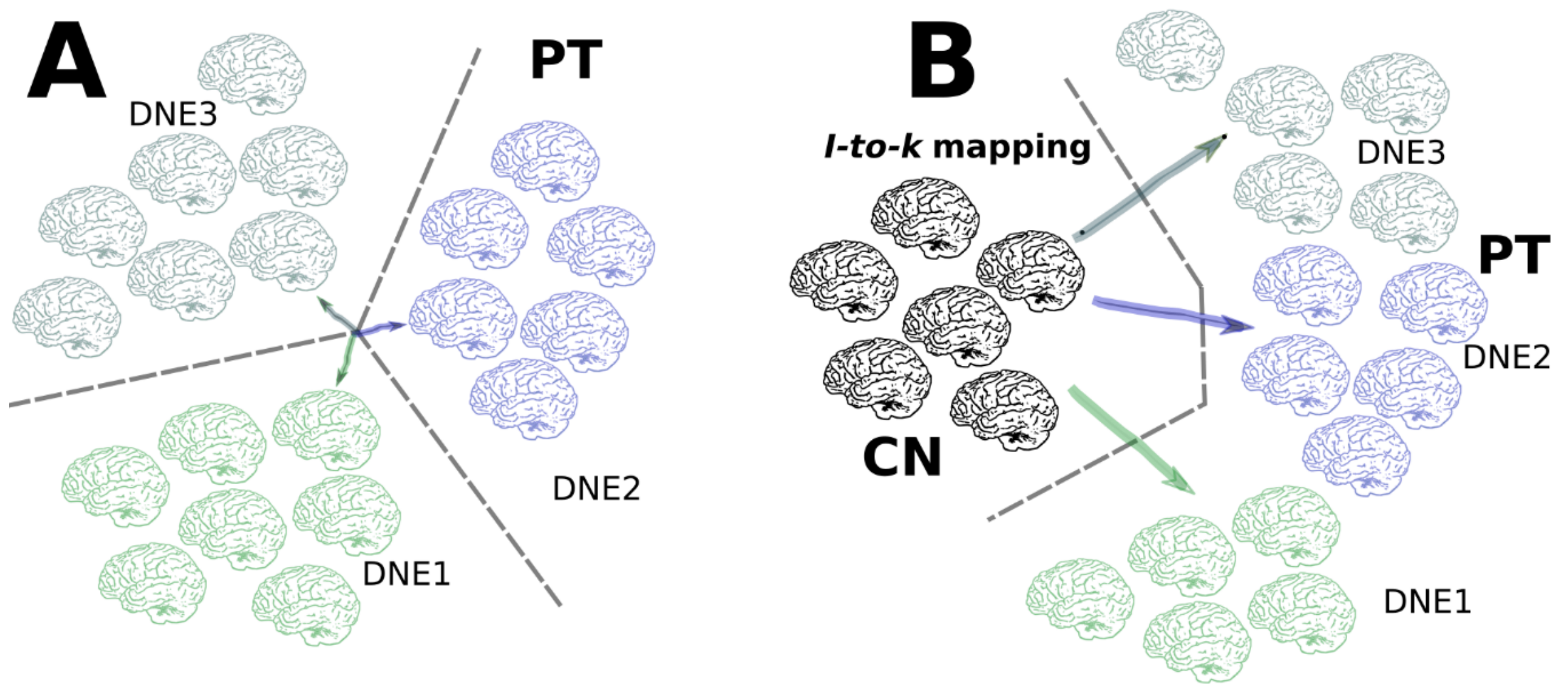

**Figure 2. Schematic diagrams of unsupervised clustering (A) and weakly-supervised clustering (B) techniques for identifying Dimensional Neuroimaging Endophenotypes (DNE)**. Unsupervised clustering methods involve performing clustering directly in the patient (PT) domain. On the other hand, weakly-supervised clustering methods aim to establish a "*1-to-k*" mapping from the healthy control (CN) to the patient (PT) domain. This approach hypothesizes that the identified dimensional neuroimaging endophenotypes (DNE) are largely driven by underlying pathological processes rather than confounding factors such as demographics.

# 3. Disease heterogeneity in Alzheimer's disease

AD, alongside its prodromal stage characterized by mild cognitive impairment (MCI), stands as one of the most prevalent neurodegenerative conditions, impacting millions worldwide[79]. Despite numerous imaging studies extracting AD-related imaging patterns[80,81], many have overlooked the neuroanatomical heterogeneity within AD.

A growing body of research has recently focused on deriving AD imaging signatures that account for this heterogeneity, recognizing subtypes, and contributing to the 'N' dimension within the ATN framework[82]. These studies largely focused on using T1w MRI and functional MRI, but recent advancements in PET also suggest variations in the distribution of tau pathology in AD[25], leading to a range of diverse syndromes. Prior research predominantly utilized unsupervised clustering techniques like SuStaIn[76] and non-negative matrix factorization[27]. Meanwhile, weakly-supervised clustering methods[14,49] were also proposed to tackle this problem. On the one hand, due to differences in databases, methodologies, and the imaging modality (MRI or PET), the reported clusters and the neuroanatomical patterns of subtypes vary and cannot be readily compared. For instance, Poulakis et al.[26] focused on dissecting the neuroanatomical heterogeneity for AD patients, while Dong et al. for AD and MCI patients[83]. However, some consistent subtypes were found across different studies. Unsurprisingly, the typical AD subtype involving the medial temporal and hippocampus was consistently found across different studies. Young et al. applied SuStaIn and found atrophy originated in the hippocampus and amygdala, which they referred to as a typical AD subtype[76]; some studies refer to this pattern as a limbic-predominant subtype. Yang et al. found this imaging pattern using the weakly-supervised clustering termed Smile-GAN[14]. Zhang et al.[34] used Bayesian latent modeling and identified the subtype of temporal lobe atrophy. Some "atypical" AD subtypes identified in the literature vary from study to study. Several neuroimaging studies have identified a subtype of AD characterized by minimal atrophy[49,84,85]. Additionally, there has been substantial observation of a subtype prominently exhibiting sparing of the hippocampus, affecting the cortex[29,84].

The clinical implications of the AD subtypes are substantial. For example, for the atypical subtypes identified above, one may ask whether these subtypes are distinct entities or comorbidity effects along the AD disease continuum. In a recent study using imaging genetics[3], we found that one dimension of AD (R1) showed widespread brain atrophy and was implicated in neurobiological processes related to cardiovascular diseases instead of typical AD pathology. Recent tau subtypes also back this finding, showing similarities to previously identified non-amnestic clinicoradiological syndromes in both early- and late-onset patients[16]. The variance in neuroanatomical heterogeneity may stem from multiple sources, encompassing genetics, environment, modifiable lifestyle factors, regional vulnerability, brain organization, and brain resilience[86]. Therefore, a thorough grasp of the entire disease spectrum relies on integrating multi-omics data, which is pivotal for future research endeavors.

# 4. Disease heterogeneity in schizophrenia

SCZ is a debilitating neuropsychiatric disorder[87] and typically clinically manifests during late adolescence or early adulthood[88]. Symptoms are grouped into three main categories: positive, negative, and cognitive, and include delusions, hallucinations, apathy, anhedonia, and memory deficits. The wide variety of symptoms leads to significant heterogeneity in the clinical presentation and clinical characteristics[89] of patients with SCZ, which likely stems from groups of patients with differing underlying neurobiological mechanisms. Consequently, patients also show significant heterogeneity in their responses to pharmacological treatments[90,91] and long-term outcomes[92]. However, there are currently no clinically available tools that can be used for diagnosis, prognosis, treatment selection, or to predict treatment response in this population. Traditional case-control neuroimaging studies are greatly affected by brain-based heterogeneity. Therefore, recent research has focused on finding more neuroanatomically homogeneous subgroups of patients using unbiased statistical techniques[93].

Most studies have focused on identifying subtypes in patients with chronic SCZ by clustering structural brain data rather than functional imaging data. The studies report between two[42,43,47,49,51], three[44,46,50], and six[48] subtype solutions. Common characteristics across subtyping studies include increased striatal volumes and frontotemporal volume reductions ranging from moderate to severe. Some solutions also implicate the insula and thalamus, all of which have been reported in large case-control meta-analytic studies[94,95].

Interestingly, multiple studies report a subtype with increased striatal volumes and few cortical effects and a second subtype with widespread cortical atrophy but few subcortical effects besides in the thalamus. The original dopamine hypothesis of SCZ implied that all patients experiencing positive symptoms have excessive striatal dopamine levels[96]. However, the apparent data-driven division of patients with and without striatal abnormalities suggests that dopaminergic dysfunction may not be present in all patients. Previous studies have shown that the pathophysiology of patients with treatment-resistant SCZ may be glutamatergic rather than dopaminergic dominant compared to patients with treatment-responsive SCZ[97,98]. Therefore, subtyping may be identifying patients who express different disease mechanisms.

Recent studies have shown that the subtypes identified in patients with chronic SCZ are also expressed at illness onset (in patients with a first episode of psychosis) and in healthy, non-clinical samples[99,100], suggesting that the subtypes represent subclinical vulnerability brain phenotypes of SCZ. The authors further found that healthy individuals who expressed one of the SCZ subtypes had higher polygenic risk scores for SCZ than those who did not express the pattern[100].

The triple-network model of SCZ implicates aberrant interactions between three key functional networks as being responsible for the array of symptoms observed in SCZ. Authors Liang et al.[45] identified two subgroups of patients by analyzing functional connections between key nodes of the networks in the triple-network model. One subtype was characterized by reduced connectivity in the salience network portion of the triple-network model. Patients in this subgroup had worse symptoms and problems with sustained attention. The second subtype exhibited hyperconnectivity of key nodes in the model, and patients had more problems with cognitive flexibility.

Taken together, the findings outlined here show that there is promise for subtyping to be used in the early stages of the disease to identify vulnerability. Identifying more homogeneous subgroups of patients with differing underlying neurobiologies could also guide future drug development and selection.

## 5. Disease heterogeneity in major depressive disorder

MDD is common and severe and affects over 320 million people worldwide[101]. A DSM diagnosis of MDD requires any 5 out of 9 symptoms to be present, resulting in a possible 227 different symptom combinations that fulfill diagnostic criteria[102,103]. Besides causing significant reductions in social and role functioning, the heterogeneous symptom profile points towards a disorder with a highly variable pathophysiology, which is also evident in the heterogeneous treatment outcomes and in the longitudinal course of the illness among patients. Currently, there are no biomarkers to aid in identifying the disorder or to predict treatment response; therefore, identifying subtypes is a step in that direction.

Recent efforts to identify more homogeneous subgroups of patients with depression have mainly focused on resting-state fMRI data, resulting in two[55,57], three[56,58], and four[54] subtypes. These subtypes are characterized by reduced connectivity in different networks, including the default mode network (DMN), ventral attention network, and frontostriatal and limbic dysfunction. Methodological discrepancies inherent to fMRI studies are highlighted in the opposing results of Drysdale et al.[52] and Liang et al.[55] in the relation between subtypes and symptoms. With sample sizes almost 10-fold larger than the other subtyping studies conducted in MDD, the two DMN-centric subtypes found by Liang et al.[55] had no relation to demographic variables or symptom severity as measured by the Hamilton Depression Rating Scale. On the other hand, disruptions in specific network components of the four subtypes identified by Drysdale et al.[52] had associations with different symptoms. For example, reduced connectivity in the front-amygdala network was associated with increased anxiety symptoms, which were most severe in subtypes 1 and 4. Hyperconnectivity in thalamic and frontostriatal networks was common in subtypes 3 and 4 and was associated with abnormal reward-driven behavior and feelings of anhedonia. Lastly, reduced connectivity in the anterior cingulate and orbitofrontal areas was most severe in subtypes 1 and 2 and was associated with problems with motivation and feelings of low energy and fatigue.

Regarding structural neuroimaging, one study used discriminative analysis on regional gray matter volumes and identified two depression subtypes in older patients. Relatively preserved gray matter volumes characterized the first subtype, and the second had widespread atrophy and white matter disruptions associated with accelerated progression to AD[57].

Although fMRI subtyping shows some promise in parsing the heterogeneity in MDD, the methodological variability across studies makes them difficult to compare and may make it harder to incorporate into clinical practice. It is unclear whether fMRI clustering is stable over time and whether patients may express different subtypes as the disorder progresses or as they experience symptom changes across different episodes. Compared to other disorders, clustering based on structural neuroimaging data is limited but should be considered in the future, along with the utility of subtypes in predicting disease outcomes.

# 6. Disease heterogeneity in autism spectrum disorder

ASD encompasses a broad spectrum of social deficits and atypical behaviors, contributing to its highly heterogeneous clinical presentation[104]. Extensive research has sought to delineate subtypes within ASD for more precise diagnostic characterization[105]. Neuroimaging studies have reported accelerated brain growth in childhood followed by a slow development into adolescence and adulthood[106]. However, these findings diverge at a localized brain level, and significant interindividual variability has been observed[107]. Initiatives such as the ABIDE (Autism Brain Imaging Data Exchange)[108] and the EU-AIMS (European Autism Interventions – A Multicentre Study)[109] have catalyzed large-scale neuroimaging subtyping projects[110].

Various clustering methods, including traditional techniques like K-means[60] or hierarchical clustering[59], have unveiled structural brain-based subtypes in ASD. Functional MRI[61] and EEG[111] are popular modalities to investigate beyond structural MRI. Given the substantial heterogeneity in the ASD population, normative clustering and dimensional analyses are deemed more suitable[63]. However, research in this area remains limited[67,112]. While validation and replication efforts are still needed to outline reliable neuroanatomical subtypes or dimensions of ASD, some convergence in structural findings is noted.

Most sets of ASD neuroimaging subtypes reveal a combination of both increases and decreases in imaging features compared to the typically developing group rather than indicating a uniform direction, highlighting the considerable heterogeneity in ASD brains. These subtypes are characterized by spatially distributed imaging patterns instead of isolated or focal patterns. Many structural MRI studies have reported widespread changes in cortical thickness[59] or brain volume[65] as key characteristics of their ASD subtypes. Functional connectivity findings have yet to converge or be replicated.

The quest for ASD subtypes faces unique challenges. Firstly, the early onset of ASD suggests a strong influence on neurodevelopmental processes, leading to potential variations in results depending on the selected age range. Secondly, ASD exhibits a higher prevalence in males, with three to four male cases for every female case[113], introducing a potential gender bias. Thirdly, individuals with ASD commonly experience psychiatric comorbidities such as ADHD, anxiety disorders, and obsessive-compulsive disorder, emphasizing the importance of careful sample selection or interpretation of the findings[114].

The DSM-5 collapsed Autistic Disorder, Asperger's Disorder, and Pervasive Developmental Disorder Not Otherwise Specified (PDD-NOS) into a single category, recognizing the continuous nature of symptoms rather than relying on distinct boundaries[115]. Aglinskas et al. argues that individuals with ASD do not cluster into distinct neuroanatomical subtypes but organize along continuous dimensions affecting specific brain regions[64]. Consequently, when designing autism models, adopting dimensional models over forcing distinct clusters is crucial. It is also imperative to contextualize ASD within the broader spectrum of mental health comorbidities[116]. Finally, as ASD is known to be highly heritable[117], the consolidation of results from various studies to formulate reproducible dimensions in the ASD model should be anchored in genetic underpinnings.

# 7. Disease heterogeneity in multiple sclerosis

MS, affecting over 2.8 million individuals worldwide, is a chronic autoimmune disorder predominantly impacting the central nervous system (CNS). In MS, the immune system targets the myelin surrounding nerve fibers, disrupting communication between the brain and the rest of the body. Over time, this condition can lead to degeneration of the nerve fibers[118]. Clinically, MS is categorized into four phenotypes based on disease activity and disability progression: clinically isolated syndrome (CIS), relapsing-remitting MS (RRMS), primary progressive MS (PPMS), and secondary progressive MS (SPMS)[119]. However, such categorization accounts solely for clinical symptoms and disregards the intricate pathobiological mechanisms underlying these symptoms/conditions, thus hindering clinical applicability.

An increasing body of literature recently used MRI to explore the heterogeneity of underpinning pathobiological mechanisms. Using the SuStain model[29], Pontillo et al.[69] detected two MRI-driven subtypes in individuals with RRMS, PPMS, and SPMS; one marked by early deep gray (GM) atrophy and lesion accrual, succeeded by cortical atrophy, and showing longer disease duration, and one characterized by cortical atrophy followed by lesion accumulation and deep GM atrophy. These findings agree with Eshaghi et al.[68], in which the subtypes were derived using the same method with slightly different features. Besides GM and lesion volumes, Eshaghi et al. also included MRI-derived measures of normal-appearing white matter (NAWM) damage in the model, and they found one more subtype exhibiting an early decrease in the T1/T2 ratio in NAWM regions, indicating widespread but subtle tissue damage, followed by GM atrophy and lesion accumulation.

Finally, Crimi et al.[70] focused on early disease stages by studying the spatiotemporal evolution of lesions in CIS patients using MRI volumes enhanced by two contrast agents highlighting different phenomena. By performing spectral clustering, they found three clusters of lesion patterns, two associated with greater total lesion volume and T1-hypointense lesions at 2-year follow-up, indicating severe and probably irreversible WM disruption correlating with anticipated future disabilities.

## 8. Disease heterogeneity from a transdiagnostic angle

The large overlap of symptoms across psychiatric disorders and the high prevalence of comorbid disorders suggests shared neurobiological processes among different psychopathologies[120]. Recent evidence shows that shared neurobiological and cellular mechanisms account for the differences in cortical thickness observed across psychiatric disorders, with strong influences from genes involved in axonal guidance and synaptic plasticity during early development[121]. Therefore, researchers have recently started investigating the neuroanatomical and neurobiological commonalities across diagnostic boundaries. A caveat is that before being able to subtype patients with different diagnoses, the patient populations are first identified and recruited using traditional nosology, which can create a vicious cycle. Nonetheless, subtyping algorithms have been applied to patients with similar groups of symptoms; for example, bipolar disorder, MDD, and SCZ share many mood-related symptoms.

Results from subtyping studies suggest that patients with different symptom-based diagnoses share transdiagnostic neuroanatomical patterns. Namely, a subset of patients with SCZ and some with bipolar disorder were grouped into a cluster that had reduced cortical thickness, cortical surface area values, and subcortical volumes, a cluster that was associated with a longer duration of illness[73]. One study also investigated commonalities in patients experiencing their first episode of depression or psychosis (rather than patients with more longstanding symptoms). Researchers found that two subtypes best characterized their sample of patients[72]. One subtype had widespread gray matter volume reductions and a more severe symptom profile, whereas the other subtype had relatively normal gray matter but increased volumes in the cerebellum compared to controls. A similar two-subtype solution characterized by widespread larger and smaller gray matter volumes and cortical thickness was also found in patients with internalizing symptoms (such as anxiety and depressive symptoms)[74]. Whilst there have yet to be larger studies encompassing and combining a larger number of diverse patient populations, including anxiety disorders, mood, thought, substance abuse, and eating disorders, the findings so far suggest that a biological-based classification system could be developed for psychiatric disorders.

## 9. Discussion

This review initiates a systematic, albeit incomplete, literature review of studies utilizing machine learning and MRI techniques to elucidate the heterogeneity of brain imaging phenotypes in various neuropsychiatric and neurodegenerative disorders. We also propose the concept of DNE for investigating disease heterogeneity, which captures dimensions of brain phenotypes associated with neurologic, neuropsychiatric, and neurodegenerative diseases. We further elaborate on three key aspects: *i*) elucidating the study paradigm of DNE that extends beyond disease subtypes, *ii*) discussing the shift from neuroimaging to multi-omics, and *iii*) highlighting its clinical implications.

The approach to unraveling disease heterogeneity using multimodal MRI through machine learning frames this problem as a clustering problem[68]. This conventional paradigm typically relies on cross-sectional data, employing machine learning algorithms to delineate distinct disease subtypes. However, this method often simplifies the intricate landscape of heterogeneous biological processes underlying the diseases. By characterizing diseases into discrete subtypes through hard-coded clustering, there is a risk of oversimplification and neglect of these conditions' dynamic and evolving nature as a continuum. Conversely, the notion of DNE aligns with the endophenotype hypothesis[20]. According to this hypothesis, DNE is envisioned to exist within the causal pathway of the disease, spanning from its underlying etiology to its exo-phenotype, such as cognitive decline. Specifically, this notion resembles the neurodevelopmental hypothesis of SCZ[122]. Disorders such as SCZ exist on a gradient of severity, suggesting that their distinctions involve both quantitative and qualitative variations to some extent. Moreover, traditional clustering approaches aimed at identifying disease subtypes may neglect the possibility that a given patient could exhibit multiple neuroanatomical patterns concurrently. However, these models often categorize patients into a singular disease subtype, overlooking the potential coexistence of multiple dimensions. In tackling this issue, representation learning[123] may offer new perspectives on understanding disease heterogeneity. In a recent study, Yang et al. introduced a weakly-supervised representation learning model named Surreal-GAN[124]. This model enables patients to exhibit brain atrophy in various distinct neuroanatomical patterns by leveraging generative adversarial networks. In subsequent applications related to Alzheimer's disease, the authors demonstrated that the derived two-dimensional scores (R1 and R2[125]) could serve as innovative instruments that can be used to establish connections between different neuroanatomical patterns and underlying genetics, providing insights into the biological mechanisms associated with each dimension. This entails a shift towards more sophisticated methodologies that go beyond cross-sectional data and incorporate temporal dynamics, recognizing the continuum of disease progression.

As stated in the endophenotype hypothesis, these DNEs are associated with underlying genetics. Recent endeavors have linked these machine learning-derived DNEs with genetics. For example, weakly-supervised machine learning methodologies[75] have been utilized in various investigations focused on late-life depression[4], autism[67], and brain aging[126]. These studies first derived the DNEs and subsequently linked these DNEs to common genetic variants. Wen et al. conducted a recent comprehensive examination of the genetic architecture underlying 9 DNEs derived from 4 common brain diseases[6]. Their study highlighted the clinical potential of these DNEs in predicting systemic disease categories. This approach facilitates a comprehensive understanding of the genetic compositions associated with the identified DNEs. An expansion of current machine learning methodologies, primarily relying on MRI data alone, could involve the incorporation of genetics[7,127–129]. This aligns with previous studies demonstrating the substantial genetic foundations contributing to disease heterogeneity. Deep learning methods[130] that jointly model imaging and genetic data can further contribute to deriving DNEs with genetic underpinnings and/or reflecting brain phenotypes associated with drug targets.

From a clinical standpoint, establishing a low-dimensional yet clinically insightful coordinate system encompassing an expanding array of DNEs can advance precision medicine[131] on multiple fronts. Firstly, breaking down disease diagnoses within a unified framework into more homogeneous dimensions can offer a more nuanced understanding of underlying neuropathological processes. This approach enables a more comprehensive capture of a particular disease's diverse brain and clinical phenotypes. By dissecting these dimensions, clinicians may gain valuable insights into the intricate factors contributing to disease manifestation. Secondly, clinical diagnoses and treatment planning can be refined by capturing the degree to which a specific DNE is expressed in an individual on a continuous scale instead of assigning them discretely to a single subtype. This continuous assessment provides a more dynamic and personalized perspective, acknowledging the variability within patient populations and tailoring interventions to individual needs. Furthermore, the precise characterization of neuropathologic phenotypes facilitated by these DNEs can substantially enhance the sensitivity of clinical trials to detect treatment effects[132]. Identifying and categorizing patients based on their unique DNE profiles allows for more targeted and efficient downstream population selections for clinical trials. This not only streamlines patient recruitment but also reduces heterogeneity in stratification, thereby optimizing the likelihood of detecting meaningful treatment effects. Integrating DNEs into clinical practice holds the promise of advancing precision medicine and refining approaches to diagnosis, treatment, and research methodologies.

## Code Availability

Machine learning methodologies using multimodal MRI to dissect disease heterogeneity:

- HYDRA: https://github.com/anbai106/mlni (Python implementation); https://github.com/evarol/HYDRA (Matlab implementation)
- CHIMERA: https://github.com/anbai106/CHIMERA (Python re-implementation)
- MAGIC: https://github.com/anbai106/MAGIC
- Smile-GAN: https://github.com/zhijian-yang/SmileGAN
- Surreal-GAN: https://github.com/zhijian-yang/SurrealGAN
- Gene-SGAN: https://github.com/zhijian-yang/GeneSGAN
- SuStaIn: https://github.com/ucl-pond/SuStaInMatlab?tab=readme-ov-file (Matlab implementation); https://github.com/ucl-pond/pySuStaIn (Python implementation)
- NMF: http://renozao.github.io/NMF/
- MCI-subtype: https://github.com/rlckd/MCI-subtype
- Bayesian LDA: https://github.com/ThomasYeoLab/CBIG/tree/master/stable_projects/disorder_subtypes/Zhang2016_ADFactors
- RCC: https://github.com/UTHSCSA-NAL/RCC-Code
- Multiple-Co-clustering: https://github.com/tomokitokuda/Multiple-Co-clustering
- pub-CVAE-MRI-ASD: https://github.com/sccnlab/pub-CVAE-MRI-ASD

## Acknowledgments

The iSTAGING consortium is a multi-institutional effort funded by NIA by RF1 AG054409 led by DC. WJ is supported by the startup funding as an assistant professor from the University of Southern California.